# Improved Trust in Human-Robot Collaboration with ChatGPT

Yang Ye, Hengxu You, and Jing Du, Ph.D.

*Abstract*— Human-robot collaboration is becoming increasingly important as robots become more involved in various aspects of human life in the era of Artificial Intelligence. However, the issue of human operators' trust in robots remains a significant concern, primarily due to the lack of adequate semantic understanding and communication between humans and robots. The emergence of Large Language Models (LLMs), such as ChatGPT, provides an opportunity to develop an interactive, communicative, and robust human-robot collaboration approach. This paper explores the impact of ChatGPT on trust in a human-robot collaboration assembly task. This study designs a robot control system called RoboGPT using ChatGPT to control a 7-degree-of-freedom robot arm to help human operators fetch, and place tools, while human operators can communicate with and control the robot arm using natural language. A human-subject experiment showed that incorporating ChatGPT in robots significantly increased trust in human-robot collaboration, which can be attributed to the robot's ability to communicate more effectively with humans. Furthermore, ChatGPT's ability to understand the nuances of human language and respond appropriately helps to build a more natural and intuitive human-robot interaction. The findings of this study have significant implications for the development of human-robot collaboration systems.

## I. INTRODUCTION

The use of robot platforms has become ubiquitous across a broad range of industries, from manufacturing and construction [1] to healthcare and education [2]. Although automation is one of the most essential advantages of robotic systems, a collaborative environment in which human operators and robotic systems work side by side is relevant due to potential task complexity, dexterity, and the need for skillful operators with specialized tools or skills. Such a Human-Robot Collaboration (HRC) work mode can lead to higher efficiency, greater accuracy, and increased productivity [3].

However, one of the most significant challenges in the HRC area is the lack of human operators' trust in robotic collaborators [4]. Trust is believed to be one of the most fundamental elements of a collaborative relationship, which impacts the willingness to cooperate and productivity [5]. Studies suggest that the lack of trust in HRC is caused by the lack of reliability, predictability, and transparency [6, 7], attributed to effective communication and mutual understanding between the human operators and the robotic system. Although previous studies have explored using natural language in robotic control [8], the robotic control models still have challenges in task context understanding, logical and procedural reasoning, and interactivity.

The recent advancement in large language models (LLMs) such as ChatGPT [9], provides a transformative opportunity for HRC. By integrating large training models [9], LLMs possess the close-to-human capability to understand the context and interact with human operators naturally. However, emerging problems associated with LLMs need to be answered. On the one hand, LLMs show great potential in improving the natural interaction between humans and robots, which can be an essential milestone for future HRC. However, the method of such integration has not been thoroughly researched. On the other hand, the trends of LLMs' integration in HRC pose unprecedented needs to understand the underlying impact on human operators, especially related to trust levels, which deserve investigation.

This study aims to explore the potential effect on trust of using an intelligent LLM for robotic control in an HRC assembly task. We proposed RoboGPT in this study. RoboGPT Integrates ChatGPT, a popular LLM, with robot control capabilities. RoboGPT retrains ChatGPT as an AI robotic control assistant by providing the task context and sample prompts. The AI assistant receives human operators' verbal commands and controls the robot arm accordingly. Such an assistant can understand the task context, procedure, and verbal commands and interact with human operators to clarify the commands or request more contextual information. We assume that the ChatGPT-enabled AI robot control assistant increases human operators' trust toward the robotic system owing to the natural bilateral communication and context understanding capabilities.

This study demonstrates and validates the HRC design using a human-subject experiment. A total of 15 participants were recruited to work with a collaborative robot via conversations in an assembly task. The collaborative robot helped human operators fetch and place tools according to

Yang Ye, Ph.D. candidate, Informatics, Cobots and Intelligent Construction Lab, Department of Civil and Coastal Engineering, University of Florida, FL 32611 USA (e-mail: ye.yang@ufl.edu).

Hengxu You, Ph.D. student, Informatics, Cobots and Intelligent Construction Lab, Department of Civil and Coastal Engineering, University of Florida, FL 32611 USA (e-mail: you.h@ufl.edu).

Jing Du, Associate Professor, Informatics, Cobots and Intelligent Construction Lab, Department of Civil and Coastal Engineering, University of Florida, FL 32611 USA (e-mail: eric.du@essie.ufl.edu). Corresponding author.

the operator's spoken language. The rest of the paper elaborates on the point of departure, the technique to build the LLM-enabled human-robot collaboration, the human-subject experiment, and the results.

## II. RELATED WORK

### A. Trust in Human-robot Collaboration

Trust is crucial in HRC, as it influences the degree to which humans are willing to work with and rely on robotic systems [10]. Trust levels can vary depending on the method of interaction in HRC, such as touchscreen control, joystick control, and verbal control [11]. To better understand this relationship, it is essential to delve deeper into the theories of trust in HRC. Several theories have been proposed to explain the differing levels of trust in human-robot collaboration. One possible reason is the lack of transparency in the robot's decision-making process. Transparency refers to the extent to which a robot's actions, intentions, and decision-making processes are clear and understandable to the human operator [12]. When humans cannot understand how a robot arrives at a specific decision or action, they may hesitate to trust it. The lack of transparency may lead to reduced trust in the robot and, consequently, a lower willingness to collaborate effectively [12].

Another potential factor contributing to trust variations in human-robot collaboration is the robotic system's ability to process and respond to ambiguous input [13]. Ambiguity of human command input can arise from various sources, such as unclear commands, conflicting information, or environmental uncertainty. Research indicates that robots capable of effectively handling ambiguous input are more likely to be trusted by human operators [14]. This is because when robots demonstrate adaptability and competence in dealing with uncertain situations, human operators gradually gain confidence in the robot's capabilities and are more willing to trust them [15]. As such, it is worth exploring whether the trust in HRC can be improved by an AI robot control assistant who can communicate with the human operator naturally and is robust to different speech styles.

### B. State-of-the-art in Human-Robot Collaboration Interface

Over the past few years, human-robot collaboration has evolved significantly, focusing on developing robots that can seamlessly interact and cooperate with humans. A notable example of HRC is using Baxter, a collaborative robot, to perform tasks such as grabbing and manipulating objects [16]. Another example is the utilization of quadruped and wheeled robots for transporting objects within warehouses and hospitals and exploring a site [17, 18]. These robots can navigate autonomously and adapt to dynamic environments, reducing the workload for human workers and increasing efficiency.

Various control methods have been employed to enhance and facilitate the interaction between humans and robots. Many of these methods involve using artificial intelligence (AI) to enable robots to understand and interpret human actions and intentions [19]. For instance, researchers have used deep learning and reinforcement learning techniques to teach robots to recognize human gestures [20, 21] and adapt to human behaviors.

The recent advancements in LLMs, such as GPTs [9], have shown the potential to improve human-robot collaboration by enabling more natural and effective communication. These models can generate human-like responses to natural language input, allowing robots to engage in more intuitive and context-aware conversations with humans [22]. However, integrating LLMs into HRC interfaces is still an emerging area of research and requires further exploration. Research on the effectiveness of these models in fostering mutual understanding and trust between humans and robots, as well as their impact on the efficiency and quality of collaboration, is essential for advancing HRC technologies.

## III. CHATGPT-ENABLED HUMAN-ROBOT COLLABORATION

### A. System Design

We establish a workflow that integrates LLMs with robotic control modules to build an intelligent AI robot control assistant, called RoboGPT. As shown in Fig 1, the RoboGPT workflow firstly transforms human operators' spoken language into textual input for the AI assistant to process. The decision-making core of the AI assistant utilizes GPT3.5 [9] in this study, to understand the information and respond. By considering the contextual information and evaluating the ambiguity of information, GPT3.5 generates natural responses to either further clarify the information with the human operators via conversations or control the robot. When communicating with human operators, the RoboGPT AI assistant generates prompts, presents the prompts to human operators, and waits for further instructions. Such bidirectional communication clarifies the intention of both human operators and the robotic system, which could increase transparency and reduce ambiguity. Then, if the RoboGPT AI assistant considers the information adequate for decision making, the responses will be sent to a decoder which further processes the commands into Robotic Operating System (ROS) topics and triggers robotic control functions to perform tasks correspondingly.

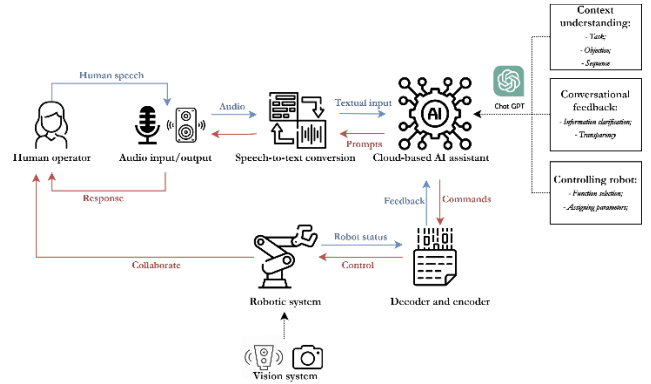

**Fig. 1.** System workflow of RoboGPT.

## B. ChatGPT Assistant

It is widely acknowledged that training an ChatGPT-enabled AI assistant from scratch is not feasible for most researchers or organizations, primarily due to the prohibitively high costs associated with data and infrastructure [23]. Given these challenges, an alternative approach that has gained considerable traction is using informative prompts to reshape and fine-tune pre-trained LLMs [24]. This method retains the capabilities embedded within existing models, allowing users to adapt the models to meet specific requirements.

In this study, we fine-tuned GPT3.5 into the RoboGPT AI robot control assistant using carefully designed prompts, as shown in Fig 2, using OpenAI API [25]. These prompts defined the system context and showed sample conversations demonstrating the desired formats. These prompts were carefully crafted to encompass various scenarios and tasks, ensuring that the fine-tuned ChatGPT could establish effective communication between human operators and robotic systems. Meanwhile, this fine-tuning process also regulated the output of the AI assistant such that the output commands could be accurately mapped with robotic control modules, which we will discuss later. After fine-tuning these messages, the GPT3.5 model could understand and process complex instructions while seamlessly integrating with various robotic platforms.

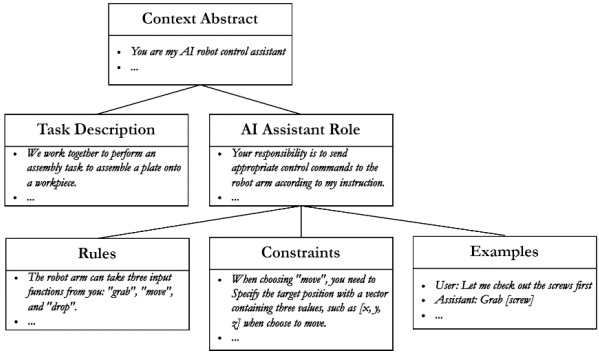

Fig. 2. Fine-tuning prompt design

## C. Robot Assembly Assistant

The robot system being tested in this study was a Franka Emika Panda robot arm [26], a lightweight seven-degree-of-freedom robot arm designed for HRC. Given that the ChatGPT controller discretely generates prompts, the target robot's posture and status change discontinuously. As such, it is necessary to map the discrete target status to a continuous trajectory for a smooth and safe robot operation. In this study, we implemented a dual-stage impedance control algorithm to resolve this issue, as shown in Figure 3. After receiving a target posture ($\tilde{x}$), the primary impedance controller generates a continuous time-series target position ($\tilde{x}_t$) to smooth the translation. The continuous target position is then fed into the secondary impedance control to generate smooth robot trajectories ($x_t$). The overall control function is described in Equation 1.

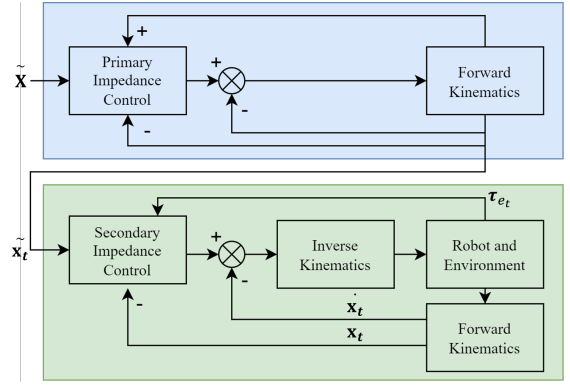

Fig. 3. Dual-stage Impedance control

$$\boldsymbol{\tau}_e = \mathbf{f}(\mathbf{g}(\tilde{x})) = \mathbf{M}\ddot{\mathbf{x}}_t + \mathbf{D}\dot{\mathbf{x}}_t + \mathbf{K}(\mathbf{x}_t - \tilde{\mathbf{x}}_t) \text{ ------ Eq (1)}$$

, where $\boldsymbol{\tau}_e$ is the external torque matrix, $\mathbf{x}_t, \dot{\mathbf{x}}_t, \ddot{\mathbf{x}}_t \in R^7$ refer to the position, velocity, and acceleration of all joints in the Cartesian space at time point t, $\tilde{\mathbf{x}}_t$ is the generated interim position in the Cartesian space at time t, $\mathbf{M}, \mathbf{D}, \mathbf{K} \in R^{7^2}$ denotes the desired inertia, damping, and stiffness respectively.

## D. Command Decoder

Then, the AI's responses need to be mapped to specific robotic control functions and variables in order to effectively harness the capabilities of the RoboGPT's ChatGPT-enabled AI assistant for robotic control. Specifically, a regular expression search algorithm [27] can be used to extract robotic control commands. This mapping process decodes the AI's responses and pulls key works to trigger robotic control algorithms with specific parameters in the ROS, subsequently initiating the execution of robot actions. In addition, the RoboGPT's AI robot control assistant can create a sequence of activities, such as grabbing an object and delivering it to a specific location directly. In such a scenario, the command decoder needs to divide the commands into several consecutive threads and execute the robot's actions sequentially.

In the context of HRC in assembly, we defined the major robotic control functions as three primary categories: *grab*, *move*, and *drop*. First, an optical input module identifies the target location when the robot needs to grab an object. Next, the robot arm executes an impedance control algorithm [28] to initiate robot arm movement while activating the robot gripper control. Finally, after a successful grabbing, the robot arm needs to deliver the object to an appropriate location before releasing the object. The AI assistant continuously keeps track of, monitors, and controls the robot's actions throughout the process.

## IV. TEST CASE AND HUMAN SUBJECT EXPERIMENT

### A. Experiment Design

To assess the impacts of ChatGPT on human operators, a human-subject experiment was conducted with 15 participants. We implemented the RoboGPT AI robot assistant to help a human operator assemble a workpiece. We selected a simple assembly scenario for this experiment to reduce the impact of the different assembly domain knowledge levels among participants. The assembly task in this experiment aims to assemble a plate onto a workpiece and fasten it with four screws using a driller. The process can be classified into three steps: assemble a plate, place screws, and drill in screws. Participants have the autonomy to decide the operation sequence, such as when to ask the robot arm to deliver tools and when to drill in screws. Human operators conduct the assembly process while the AI assistant controls the robot arm to deliver tools. The experiment has been approved by the university Institutional Review Board (IRB202300667).

This experiment was conducted in Virtual Reality (VR) to record participants' behavior and cognitive data as well as ensure participants' safety. Meanwhile, the AI assistant controlled a real 7-degree-of-freedom robot arm, Franka Emika [26], was controlled by the AI assistant. In addition, a digital twin of the real robot arm was constructed in the VR environment using our previously developed methods [1, 29]. This robot arm digital twin synchronized the robot status in VR and provided real robotic behavior to boost participants' sense of immersion.

This experiment compared two conditions. First, the fixed verbal condition served as the control condition, in which the participants interacted with the robot arm using fixed spoken language, such as "*grab the ruler*", "*move forward*", and "*drop-down*". In this condition, the robot arm only responded to exact verbal commands and performed a pre-defined fixed behavior. Another condition was the ChatGPT-enabled condition which controlled the robot arm using the RoboGPT's AI assistant. In this condition, participants were allowed to instruct the robot in any natural way they preferred, and the robot arm controller could understand different ways of speech and communicate with the robot arm bilaterally.

### B. Procedure and Data Collection

After consenting and collecting demographic information, participants were asked to go through a training session in which they practiced the assembly process with all the tools already placed in the reachable position and no need to work with the robot. Upon successfully finishing the task, participants were asked to perform the assembly task with the robot. To reduce the impact of individual differences, a within-subject design was adopted: participants needed to go through both conditions in a randomized order. After each condition, participants were required to complete a trust scale [30] and a NASA TLX [31] questionnaire. At the end of the experiment, an interview section was conducted for every participant to collect opinions regarding this HRC experiment.

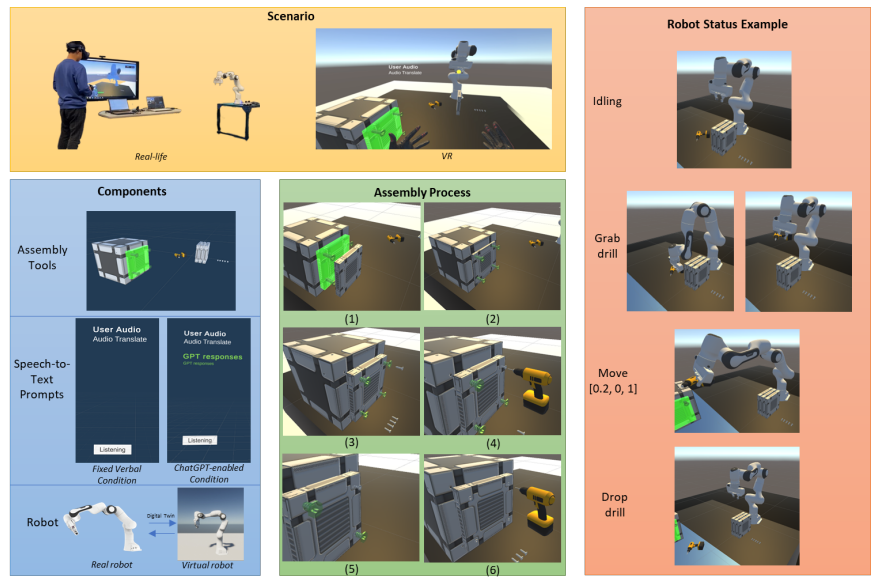

**Fig. 4.** Experiment scenario and task

## V. RESULTS

### A. Interactions

Figure 5 shows an example of a conversation in this experiment while working with the robot. Although the user speech-to-text module mistakenly converted the speech in line 3 and 15 (misinterpreted "*screw*" and "*driller*" as "*school*" and "*jeweler*"), the RoboGPT's AI assistant could tolerate such errors and make correct decisions. Meanwhile, the vague expression in Line 17 was clarified before executing robot commands.

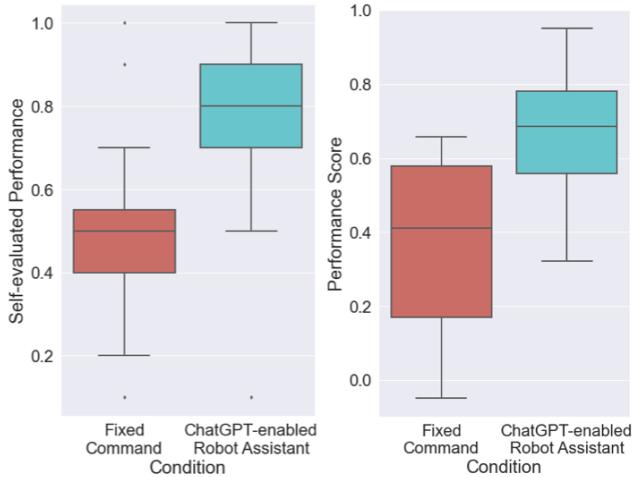

**Fig. 5.** Interaction prompt examples

*B. Performance*

We measured the assembly task performance using two matrices: performance score as indicated by completion time, and self-evaluated performance as indicated by the questionnaire. The performance score is calculated using the equation below:

$$S_i = \frac{t_{max} - t_i}{t_{max} - t_{min}} \quad \text{Eq(2)}$$

, where $t_{max}$ and $t_{min}$ refer to the maximum and minimum time participants used to finish the task. $S_i$ and $t_i$ refer to the performance score and completion time of the participant $i$. The higher the calculated performance score, the quicker the participant completed the task and the better the performance.

We also extracted participants' performance from their self-evaluation in the NASA TLX questionnaire. Both the objective performance score and the subjective self-evaluation results are shown in Figure 6.

**Fig. 6.** Task performance result

Both the performance score and self-evaluated performance results passed the Anderson Darling [32] normality test. T-test showed significant differences between the fixed command condition and the RoboGPT's AI assistant condition, with p < 0.001 for both performance matrices. These results indicate that using the LLM-enabled robot assistant could effectively increase the HRC task performance compared with fixed commands. Participants in the post-experiment interview reported that a potential reason could be that the ChatGPT-enabled robot assistant could memorize previous patterns and thus accelerate the task progress.

*C. Trust and Cognitive Load*

Trust and cognitive load were measured using the questionnaire responses. Figure 7 visualizes the results. Like the previous statistical analysis pipeline, these matrices showed significant differences between the two conditions, with p < 0.001 for both trust and cognitive load. These results indicated that participants perceived less mental load while working with the ChatGPT-enabled robot assistant. Meanwhile, they generally trusted the robot more if it could react to natural communication.

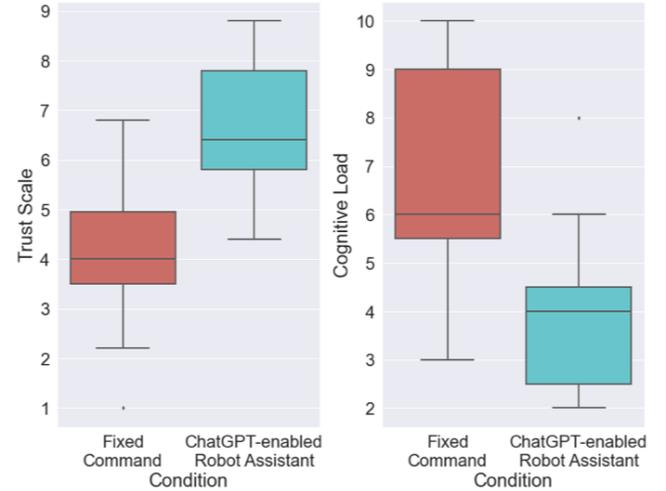

**Fig. 7.** Cognitive load and trust scale results

## VI. DISCUSSION AND LIMITATIONS

This study discusses and demonstrates the feasibility of humans using the emerging ChatGPT to collaborate with a robot arm intuitively and effectively. ChatGPT can be fine-tuned into intelligent robot assistants which can understand the task context and generate ROS messages to control a robot arm.

The experiment in this study shows that the ChatGPT-enabled robot assistant improves task performance and is believed to be more trustworthy compared to a fixed or pre-defined control method. According to the post-experiment interview, the enhanced trust can be attributed to the superior competence in completing the HRC assembly task. Furthermore, the ChatGPT-enabled robot assistant communicates with the human operator and retains memory from previous decisions, such as the location to deliver the tools. These natural ways of interaction foster better partner relationships and improve trust in this HRC task.

Furthermore, participants generally believed that working with a ChatGPT-enabled robot assistant was less mentally demanding, potentially due to its autonomy and intelligence, which reduced the attention needed to control the robot.

However, we found that the ChatGPT-enabled robot assistant was negatively evaluated in several aspects. For instance, the ChatGPT -enabled assistant could sometimes be self-assertion. The intelligence and scenario-understanding capabilities of ChatGPT -enabled assistants are double-edged swords. On the one hand, these capabilities boosted trust and reduced mental effort during the HRC. On the other hand, the ChatGPT-enabled assistants made decisions based on their understanding, which could sometimes be problematic, especially in the case of miscommunication or inaccurate communication. Furthermore, the I/O was in text format, which imposed a burden on effective communication. In real-world operations, the text is typically not adequate to describe the working scenario, such as the location of objects, the status of the robot arm, or occurred events (e.g., collision, operation failure). Therefore, the robot assistant should optimally be capable of understanding different modalities, such as imaginary inputs, to achieve better HRC results.

## VII. Conclusions

We have proposed a novel design approach for HRC using ChatGPT assistant by fine-tuning the ChatGPT to understand task context and directly control ROS topics corresponding to the human operator's verbal commands. The design was demonstrated with an HRC assembly task. We also performed a human-subject experiment in which participants reported higher trust and lower cognitive load compared with using fixed control commands. The experiment results implied a significant impact of utilizing the emerging LLMs, such as ChatGPT, to work and collaborate with robots.

In future research, it is recommended to investigate an effective HRC workflow using LLMs, such as an endorsement mechanism, before sending commands to ROS. A proper HRC workflow could maximize the benefits of LLMs as well as create comfortable and intuitive working environments for human operators.

## VIII. Acknowledgments

The authors would like to thank all the experiment participants for joining this study and contributing their thoughts. This material is supported by the National Science Foundation (NSF) Grant 2024784.

## References

[1] T. Zhou, Q. Zhu, and J. Du, "Intuitive robot teleoperation for civil engineering operations with virtual reality and deep learning scene reconstruction," *Advanced Engineering Informatics,* vol. 46, p. 101170, 2020.

[2] H. S. Ahn, M. H. Lee, and B. A. MacDonald, "Healthcare robot systems for a hospital environment: CareBot and ReceptionBot," in *2015 24th IEEE international symposium on robot and human interactive communication (RO-MAN)*, 2015: IEEE, pp. 571-576.

[3] A. Ajoudani, A. M. Zanchettin, S. Ivaldi, A. Albu-Schäffer, K. Kosuge, and O. Khatib, "Progress and prospects of the human–robot collaboration," *Autonomous Robots,* vol. 42, pp. 957-975, 2018.

[4] G. Charalambous, S. Fletcher, and P. Webb, "The development of a scale to evaluate trust in industrial human-robot collaboration," *International Journal of Social Robotics,* vol. 8, pp. 193-209, 2016.

[5] J. D. Lee and K. A. See, "Trust in automation: Designing for appropriate reliance," *Human factors,* vol. 46, no. 1, pp. 50-80, 2004.

[6] P. A. Hancock, D. R. Billings, K. E. Oleson, J. Y. Chen, E. De Visser, and R. Parasuraman, "A meta-analysis of factors influencing the development of human-robot trust," 2011.

[7] S. Ososky, D. Schuster, E. Phillips, and F. G. Jentsch, "Building appropriate trust in human-robot teams," in *2013 AAAI spring symposium series*, 2013.

[8] C. Matuszek, E. Herbst, L. Zettlemoyer, and D. Fox, "Learning to parse natural language commands to a robot control system," in *Experimental robotics: the 13th international symposium on experimental robotics*, 2013: Springer, pp. 403-415.

[9] T. Eloundou, S. Manning, P. Mishkin, and D. Rock, "Gpts are gpts: An early look at the labor market impact potential of large language models," *arXiv preprint arXiv:2303.10130,* 2023.

[10] I. Maurtua, A. Ibarguren, J. Kildal, L. Susperregi, and B. Sierra, "Human–robot collaboration in industrial applications: Safety, interaction and trust," *International Journal of Advanced Robotic Systems,* vol. 14, no. 4, p. 1729881417716010, 2017.

[11] B. Sadrfaridpour, H. Saeidi, J. Burke, K. Madathil, and Y. Wang, "Modeling and control of trust in human-robot collaborative manufacturing," *Robust intelligence and trust in autonomous systems,* pp. 115-141, 2016.

[12] R. H. Wortham and A. Theodorou, "Robot transparency, trust and utility," *Connection Science,* vol. 29, no. 3, pp. 242-248, 2017.

[13] C. Benn and A. Grastien, "Reducing moral ambiguity in partially observed human–robot interactions," *Advanced Robotics,* vol. 35, no. 9, pp. 537-552, 2021.

[14] D. Cameron *et al.*, "Framing factors: The importance of context and the individual in understanding trust in human-robot interaction," 2015.

[15] K. E. Schaefer, S. G. Hill, and F. G. Jentsch, "Trust in human-autonomy teaming: A review of trust research from the us army research laboratory robotics collaborative technology alliance," in *Advances in Human Factors in Robots and Unmanned Systems: Proceedings of the AHFE 2018 International Conference on Human Factors in Robots and Unmanned Systems, July 21-25, 2018, Loews Sapphire Falls Resort at Universal Studios,*

*Orlando, Florida, USA 9*, 2019: Springer, pp. 102-114.

[16] K. S. Haring *et al.*, "Robot authority in human-robot teaming: Effects of human-likeness and physical embodiment on compliance," *Frontiers in Psychology,* vol. 12, p. 625713, 2021.

[17] K. Talamadupula, J. Benton, S. Kambhampati, P. Schermerhorn, and M. Scheutz, "Planning for human-robot teaming in open worlds," *ACM Transactions on Intelligent Systems and Technology (TIST),* vol. 1, no. 2, pp. 1-24, 2010.

[18] P. Xia, F. Xu, T. Zhou, and J. Du, "Benchmarking Human versus Robot Performance in Emergency Structural Inspection," *Journal of Construction Engineering and Management,* vol. 148, no. 8, p. 04022070, 2022.

[19] A. Tabrez, M. B. Luebbers, and B. Hayes, "A survey of mental modeling techniques in human–robot teaming," *Current Robotics Reports,* vol. 1, pp. 259-267, 2020.

[20] T. Iqbal and L. D. Riek, "Human-robot teaming: Approaches from joint action and dynamical systems," *Humanoid robotics: A reference,* pp. 2293-2312, 2019.

[21] D. Hadfield-Menell, S. J. Russell, P. Abbeel, and A. Dragan, "Cooperative inverse reinforcement learning," *Advances in neural information processing systems,* vol. 29, 2016.

[22] M. Aljanabi, "ChatGPT: Future directions and open possibilities," *Mesopotamian Journal of CyberSecurity,* vol. 2023, pp. 16-17, 2023.

[23] A. Muhamed *et al.*, "CTR-BERT: Cost-effective knowledge distillation for billion-parameter teacher models," in *NeurIPS Efficient Natural Language and Speech Processing Workshop*, 2021.

[24] H. Liu *et al.*, "Few-shot parameter-efficient fine-tuning is better and cheaper than in-context learning," *Advances in Neural Information Processing Systems,* vol. 35, pp. 1950-1965, 2022.

[25] OpenAI. "OpenAI API reference." https://platform.openai.com/docs/api-reference (accessed 03/13, 2023).

[26] FrankaEmika. "FrankaEmika Robot Arm." https://www.franka.de/ (accessed.

[27] K. Thompson, "Programming techniques: Regular expression search algorithm," *Communications of the ACM,* vol. 11, no. 6, pp. 419-422, 1968.

[28] N. Hogan, "Impedance control: An approach to manipulation," in *1984 American control conference*, 1984: IEEE, pp. 304-313.

[29] Y. Ye, T. Zhou, and J. Du, "Robot-Assisted Immersive Kinematic Experience Transfer for Welding Training," *Journal of Computing in Civil Engineering,* vol. 37, no. 2, p. 04023002, 2023/03/01 2023, doi: 10.1061/JCCEE5.CPENG-5138.

[30] S. M. Merritt, "Affective processes in human–automation interactions," *Human Factors,* vol. 53, no. 4, pp. 356-370, 2011.

[31] S. G. Hart, "NASA-task load index (NASA-TLX); 20 years later," in *Proceedings of the human factors and ergonomics society annual meeting*, 2006, vol. 50, no. 9: Sage publications Sage CA: Los Angeles, CA, pp. 904-908.

[32] F. W. Scholz and M. A. Stephens, "K-sample Anderson–Darling tests," *Journal of the American Statistical Association,* vol. 82, no. 399, pp. 918-924, 1987.